\pdfoutput=1

\documentclass[11pt]{article}

\usepackage{ACL2023}

\usepackage{times}
\usepackage{latexsym}

\usepackage[T1]{fontenc}

\usepackage[utf8]{inputenc}

\usepackage{microtype}

\usepackage{inconsolata}

\usepackage{multirow}
\usepackage{graphicx}
\usepackage{amsmath}
\usepackage{amssymb}
\usepackage{multirow}
\usepackage{enumitem}
\usepackage{array}
\usepackage{xcolor,colortbl}
\definecolor{todo}{rgb}{1,0.5,0}

\definecolor{done}{rgb}{0.6,0,0.6}

\usepackage{float}

\definecolor{new}{rgb}{0,0,1}

\title{Systematic Task Exploration with LLMs:\\A Study in Citation Text Generation}

\author{Furkan \c{S}ahinu\c{c}$^{1}$, Ilia Kuznetsov$^{1}$, Yufang Hou$^{2,3}$, Iryna Gurevych$^{1}$ \\
$^{1}$Ubiquitous Knowledge Processing Lab (UKP Lab) \\
Department of Computer Science and Hessian Center for AI (hessian.AI) \\
Technical University of Darmstadt \\
$^{2}$IBM Research Europe - Ireland \\
$^{3}$Technical University of Darmstadt \\
\url{www.ukp.tu-darmstadt.de}}

\begin{document}
\maketitle
\begin{abstract}
Large language models (LLMs) bring unprecedented flexibility in defining and executing complex, creative natural language generation (NLG) tasks. Yet, this flexibility brings new challenges, as it introduces new degrees of freedom in formulating the task inputs and instructions and in evaluating model performance. To facilitate the exploration of creative NLG tasks, we propose a three-component research framework that consists of systematic input manipulation, reference data, and output measurement. We use this framework to explore citation text generation -- a popular scholarly NLP task that lacks consensus on the task definition and evaluation metric and has not yet been tackled within the LLM paradigm. Our results highlight the importance of systematically investigating both task instruction and input configuration when prompting LLMs, and reveal non-trivial relationships between different evaluation metrics used for citation text generation. Additional human generation and human evaluation experiments provide new qualitative insights into the task to guide future research in citation text generation. We make our code\footnote{GitHub: \href{https://github.com/UKPLab/acl2024-citation-text-generation}{UKPLab/acl2024-citation-text-generation}} and data\footnote{Data: \href{https://tudatalib.ulb.tu-darmstadt.de/handle/tudatalib/4234}{TUdatalib}} publicly available.
\end{abstract}

\section{Introduction}\label{sec:intro}

Thanks to their instruction-following abilities, large language models (LLMs) allow specifying and executing NLP tasks with unprecedented flexibility and speed, while reducing the need for task-specific architecture design, data annotation, and model training \cite{Touvron:2023-llama,Touvron:2023-llama-2,Taori:2023,Ouyang:2022,Openai:2023,Chung:2022}. This has led to a surge of new, complex, creative natural language generation (NLG) tasks like peer review generation \cite{robertson2023gpt4} or story and poetry generation \cite{Chakrabarty:2023}, that push the boundary of what was deemed feasible for NLP systems just a few years ago.

\begin{figure}[t]
  \centering
  \includegraphics[width=0.83\columnwidth]{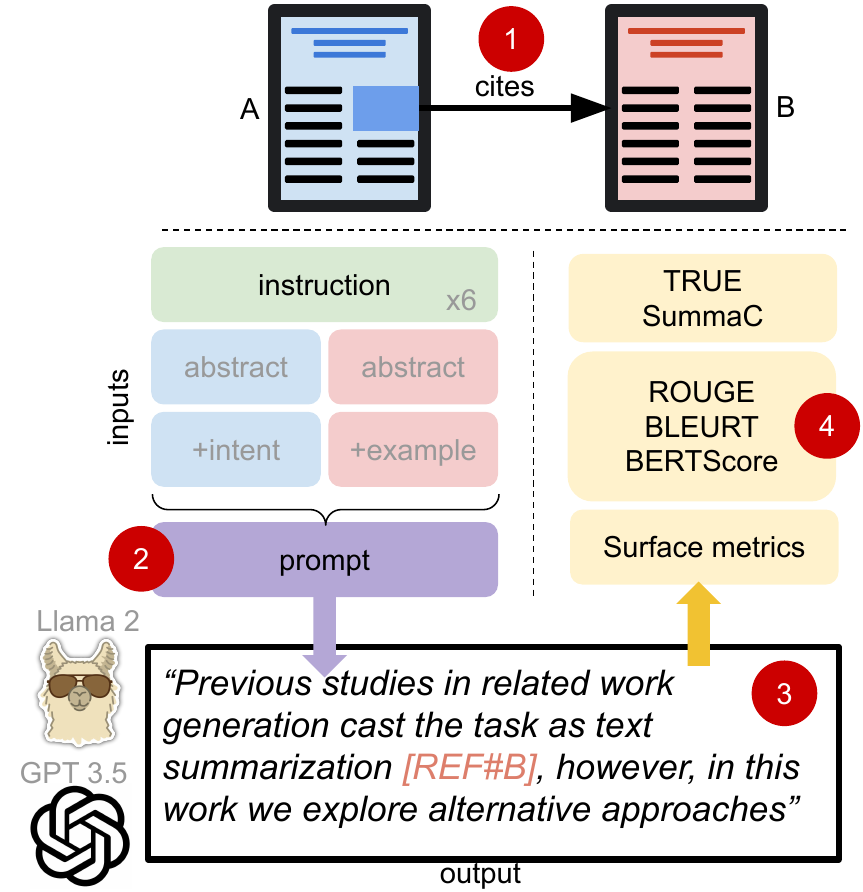}
  \caption{Citation text generation with LLMs. The task~(1) is to generate a paragraph of related work from the citing paper (A) about a cited paper (B). The instruction combined with task inputs constitutes a prompt (2) that is communicated to the model. The model's response (3) is evaluated using a range of measurements, from word count to NLI-based factuality metrics (4).}
  \label{fig:task_overview}
\end{figure}

\begin{figure*}[ht]
  \centering
  \includegraphics[width=0.84\textwidth]{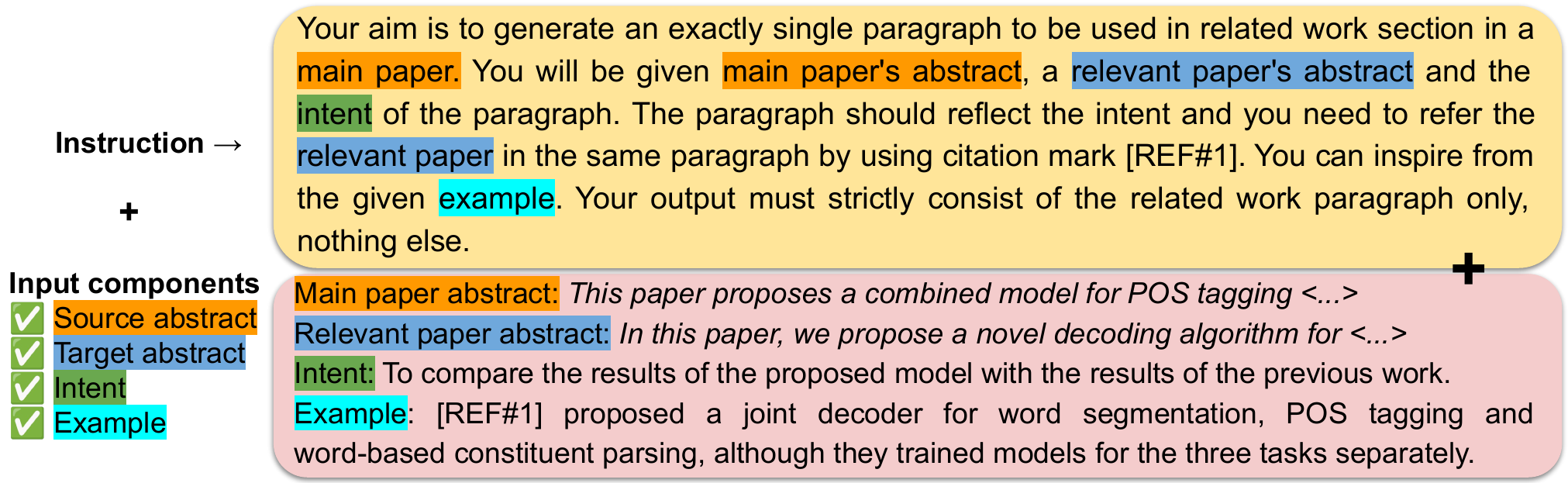}
  \caption{Prompt manipulation combines the instruction (top) with input components (left) and the corresponding data (bottom) incl. free-form citation intent and example sentence. The result serves as LLM prompt as in Figure \ref{fig:task_overview}.}
  \label{fig:prompting}
\end{figure*}

The flexibility comes at a cost, as it introduces new degrees of freedom into the analysis. LLMs generate \emph{output} in response to a \emph{prompt}, which consists of a natural-language \emph{task instruction} supplemented by additional bits of information about an instance, which we term \emph{input components} (Figure \ref{fig:prompting}). LLM-powered creative NLG tasks often feature a complex input component space, and the task instruction wording can affect model behavior in non-intuitive ways. The \emph{output space} is varied as well, as there might exist infinitely many acceptable generations. This overall variability brings the risk of creative NLG tasks being defined and evaluated ad hoc, hindering systematic comparison of NLP systems and leading to anecdotal accounts of LLM capabilities.

Although optimizing model instructions to maximize performance of LLMs is an active research area (Section \ref{sec:llms}), prompt engineering mostly targets the tasks where input and output spaces are well-defined (e.g., question answering). However, some creative NLG tasks need a step of exploration of what inputs are required and how the evaluation of outputs will be carried out before deeply exploring the best way to introduce the task to LLMs.

Our work addresses task variability in \textbf{citation text generation} -- a widely studied scholarly NLG task aiming to increase efficiency of scientific work \cite{Li:2022-automatic,Funkquist:2023}. Citation text generation is a good example of creative NLG, as it features a complex input component space combined with multiple plausible outputs. Prior work on citation text generation lacks consensus on the required inputs, explores only a limited number of measurements to characterize the outputs, and does not investigate the use of instruction-tuned LLMs to tackle the task (Table \ref{tab:rw_table}).

To address this gap, we design a framework to systematically explore the task of citation text generation with LLMs (Figure \ref{fig:task_overview}). We systematically manipulate the input components and instructions communicated to the model via a prompt, and study the effects of these manipulations on the model output using a wide range of measurements, supplemented by a novel reference dataset for citation text generation based on the ACL Anthology, and featuring novel use of free-form citation intents to guide generation (Section \ref{sec:method}). Our experiments with two state-of-the-art LLMs -- Llama 2-Chat \cite{Touvron:2023-llama-2} and GPT 3.5 Turbo \cite{Ouyang:2022} reveal that input components and task instructions both impact the generations, and their effects add up. Free-form citation intents, as illustrated in Figure \ref{fig:prompting}, show promise as an alternative to categorical intents used in prior citation text generation work. Our results (Section \ref{sec:results}) imply that the \emph{relative} performance of alternative task input configurations can be estimated on a small set of instructions, while the best \emph{absolute} performance needs experimentation with a wide array of instruction wordings. Through correlation analysis, we observe that the NLG metrics in our measurements are complementary, motivating the use of wide-spanning measurement sets for NLG tasks beyond citation text generation. Our human studies (Section \ref{sec:humeval}) reveal both quantitative and qualitative insights about input components and task instructions from both generation and evaluation perspectives.

In summary, this work contributes:
\begin{itemize}[leftmargin=*,noitemsep,topsep=0pt]
    \item A framework for exploring the task of citation text generation with LLMs;
    \item A new reference corpus of citation texts based on the ACL Anthology enriched with novel free-form citation intents; 
    \item Experimental results on the impact of task inputs and instructions on citation text generation outputs, and an examination of the relationships between the measurements;
    \item Human evaluation and generation studies providing additional insights to shape future work in citation text generation and creative NLG.
\end{itemize}

We stress that our work neither seeks nor claims state-of-the-art citation text generation, as the differences in pre-trained model capabilities would hinder a fair comparison and likely lead to confounding \cite{Nityasya:2023}. Instead, the objective of our work is to explore \textit{prompting as a tool for systematic task manipulation} in the LLM age. We believe our approach to be general and adaptable to other creative NLG tasks.

\section{Background}\label{sec:related_work}

\subsection{LLMs and Prompting}\label{sec:llms}

Instruction-tuned LLMs demonstrate competitive zero-shot performance across a wide range of NLP tasks
\cite{Touvron:2023-llama,Touvron:2023-llama-2,Taori:2023,Ouyang:2022,Openai:2023,Chung:2022}. Unlike traditional models, LLMs can be prompted with free-form textual queries -- prompts. Prompts can be manipulated through simple textual adjustments, allowing the user to guide model behavior at inference time without updating the model. Arriving at an optimal prompt is not trivial. \citet{Karmaker:2023} highlight the difficulties of prompting for complex NLG tasks and propose a taxonomy for prompt designs to facilitate NLP system comparison. Current LLMs are known to be sensitive to minor changes in task wording \cite{Brown:2020,Kojima:2022,Sanh:2022, Lu:2022,Mishra:2022,Wang:2023-robustness,Zhu:2023}, and several methods to arrive at an optimal task wording have been proposed \cite{Gonen:2023,Yin:2023,Gu:2023-robustness,Lou:2023}. In-context learning based on task demonstrations has also shown promise \cite{Ouyang:2022,Wang:2022-super-natural,Wang:2023-self-Instruct, Chung:2022}, and attracted further critical scrutiny \cite{Min:2022}. Contributing to this line of research, in addition to investigating how to introduce tasks to the models, we study the impact of alternative \textit{input configurations} on LLM behavior, exemplified by citation text generation.

\begin{table*}[ht]
\centering
\small
\renewcommand{\arraystretch}{1.0}
\begin{tabular}{l||c|c|c|c|c|l}
\hline
\textbf{Study} & \textbf{Level} & \textbf{Abstract} & \textbf{Intent} & \textbf{Example} & \textbf{Model} & \textbf{Evaluation} \\
\hline
\cite{AbuRaed:2020} & sent & Tgt & - & - & PG & ROUGE \\
\hline
\cite{Xing:2020} & sent & Tgt & - & - & PG & ROUGE, Human \\
\hline
\cite{Ge:2021} & sent & Tgt & C & - & Enc. + LSTM & ROUGE, Human \\
\hline
\cite{Kasanishi:2023} & para & Tgt & - & - & FiD & ROUGE, Human \\
\hline
\cite{Chen:2021} & para & Tgt & - & - & Hier. Enc. & ROUGE, Human \\
\hline
\cite{Li:2022-corwa} & span & Tgt & - & - & LED & ROUGE, Human \\
\hline
\cite{Luu:2021} & sent & Src/Tgt & - & - & GPT-2 & ROUGE, BLEU, Human  \\
\hline
\cite{Lu:2020} & para & Src/Tgt & - & - & PG & ROUGE, Human \\
\hline
\cite{Arita:2022} & sent & Src/Tgt & C & - & T5 & ROUGE \\
\hline
\cite{Jung:2022} & sent & Src/Tgt & C & - & T5, BART & ROUGE, SciBERTScore, Human \\
\hline
\cite{Wu:2021} & para & Src/Tgt & C & - & FiD & ROUGE, BLEU, BLEURT, Meteor \\ 
\hline \hline
\multirow{2}{*}{Ours}  & \multirow{2}{*}{para} & \multirow{2}{*}{Src/Tgt} & \multirow{2}{*}{F} & \multirow{2}{*}{\checkmark} & Llama 2-Chat & ROUGE, (Sci)BERTScore, BLEURT \\ 
& & & & & GPT-3.5 & Surface, TRUE, SummaC, Human\\
\hline
\end{tabular}
\caption{Our and prior work on citation text generation. We explore alternative task input configurations for citation text generation in the context of state-of-the-art instruction-following LLMs, using a comprehensive measurement kit and two novel input components: free-form citation intent and example sentence. sent -- sentence, para -- paragraph, span -- span of several tokens, PG -- pointer-generator network, FiD -- fusion-in-decoder network, C -- categorical intents, F - free-form intents, Src - source (citing) paper, Tgt - target (cited) paper.}
 \label{tab:rw_table}
\end{table*}

\subsection{Citation Text Generation} 

Citation text generation is a widely studied task aiming to increase the efficiency of scientific work. It has been cast as a sentence-level  \cite{AbuRaed:2020,Ge:2021,Li:2022-corwa,Li:2023} and paragraph-level task \cite{Lu:2020,Chen:2021, Chen:2022,Wu:2021,Kasanishi:2023}, as extractive \cite{Hoang:2010,Hu:2014,Chen:2019,Wang:2020} and abstractive summarization \cite{AbuRaed:2020,Li:2022,Lu:2020,Chen:2021,Luu:2021,Kasanishi:2023}. Different input components such as categorical citation intents and citation network information have been explored   \cite{Wu:2021,Arita:2022,Gu:2022,Jung:2022,Ge:2021,Wang:2021,Wang:2022,Chen:2022,Gu:2023}. Table \ref{tab:rw_table} summarizes task definitions and modeling approaches from prior work: we are the first to systematically assess the impact of different task input configurations and instructions for citation text generation using LLMs.

In addition, we explore the impact of citation intents on citation text generation. Citation intent prediction \cite{Teufel:2006,Abu-jbara:2013,Jurgens:2018,Cohan:2019,Lauscher:2022} and the use of intent in generating citation text \cite{Wu:2021,Arita:2022,Gu:2022,Jung:2022} have been previously investigated using categorical citation intents, which have a potential drawback of being not informative enough to steer the generation. To tackle this challenge, we propose novel free-form citation intents (Section \ref{sec:data}), evaluate their effects on citation text generation, and discuss the advantages and potential pitfalls of this new approach.

\subsection{NLG Evaluation}
Natural language generation is notoriously hard to evaluate automatically \cite{Gehrmann:2023}, and human evaluation is often associated with high cost and low reproducibility \cite{Belz:2023}. Conventional automatic evaluation metrics based on token or token embedding similarity like ROUGE \cite{Lin:2004}, BERTScore \cite{Zhang:2020}, or BLEURT \cite{Sellam:2020} are widely used in NLG. Yet, these metrics cannot detect factual errors in the model output, and do not capture whether the generated texts meet the formal requirements that the task imposes on the output.

To address the lack of factuality evaluation, metrics based on natural language inference (NLI) can be employed, e.g. TRUE \cite{Honovich:2022} and SummaC \cite{Laban:2022} aim to detect compatibility between the generated output and the reference. Formal evaluation of the outputs can be addressed by using surface-level measurements to check whether task instructions are followed -- yet this type of analysis is often omitted \cite{Jang:2022}. While prior work in citation text generation mostly relies on ROUGE (Table \ref{tab:rw_table}), our measurements encompass conventional, surface-level, and NLI-based metrics and enable comprehensive analysis of the generated texts and the relationships between the metrics. We complement our measurements by human evaluation, in which we both study non-author human-generated citation texts via automatic metrics, and manually evaluate machine-generated citation texts against the gold reference (Section \ref{sec:humeval}).

\section{Task and Method} \label{sec:method}

To recap, we aim to explore the impact of task input configuration and instructions on citation text generation outputs in the context of state-of-the-art LLMs. We focus on \textbf{paragraph-level generation for the related work section paragraphs}, as this represents the dominant use case for citation text generation \cite{Lu:2020,Wu:2021,Chen:2021,Kasanishi:2023}. The key components of our framework are prompt manipulation, reference data, and the measurement kit.

\subsection{Prompt Manipulation} \label{sec:prompt}

Prompt manipulation consists of systematic variation of input components and the subsequent task instructions. We experiment with four types of input components, combined with six distinct dynamically-adjusted human-written task instructions. The input components investigated are:

\begin{itemize}[leftmargin=*,noitemsep,nolistsep]
    \item \textbf{Target (cited) paper abstract:} The abstract of the cited paper is expected to contain core information about the cited work.
    
    \item \textbf{Source (citing) paper abstract:} The abstract of the citing paper is expected to provide additional context to guide generation. Cited and citing paper abstracts are commonly used in citation text generation literature (Table \ref{tab:rw_table}).
    
    \item \textbf{Citation intent:} Indicates the aim of the citation. We explore two kinds of citation intents derived from the reference paragraph: categorical intents ("\textit{Methods}") and novel free-form intents ("\textit{To compare the methods to prior work}"), discussed at length below.
    
    \item \textbf{Example sentence:} An example sentence that refers to the cited paper but does not belong to the currently considered citing paper. This input component aims to demonstrate how the paper-to-be-cited has been contextualized in other papers and serves as a proxy for a full in-context-learning example.
\end{itemize}

We use the chosen configuration of input components to generate dynamically adjusted task instructions based on six human-written instruction templates. The templates are diverse and represent different prompting techniques such as direct instruction, chain-of-thought, role-playing, and instruction list (see Appendix \ref{app:prompts}). The final prompt passed to the model is constructed by adjusting the instruction based on the chosen input component combination and concatenating the instruction with the input data for a given instance. Figure \ref{fig:prompting} provides an example: the instruction (Template~1 in Appendix \ref{app:prompts}) requests the model to compose a single related work paragraph based on the input components from the citing and cited papers, while using \texttt{[REF\#1]} to refer to the cited paper.

\subsection{Reference Data}\label{sec:data}

\textbf{Requirements.} To explore the space of possible task inputs, our study requires rich reference data. For paragraph-level generation, the data must contain \textit{full paragraphs}. We further focus on paragraphs that belong to \textit{related work sections}, where the authors are most likely to discuss cited work rather than their own contributions, compared to other sections. This requires the papers to be \textit{structured at least on the section level}. The cited papers' data should be \textit{readily accessible} based on the citation. Both citing and cited papers should be complemented with \textit{metadata}, including at least their abstracts, since this information is commonly used to generate citation texts. Among public datasets, \citet{Kasanishi:2023} and \citet{Lu:2020} come closest to our requirements. Yet, \citet{Kasanishi:2023} is limited to the literature review and survey papers, and our preliminary investigation of \citet{Lu:2020} has shown that some abstracts and citations were missing from the data. 

\textbf{Dataset.} To address these limitations, we compiled a new reference dataset based on the parsed ACL Anthology by \citet{Rohatgi:2022}. The dataset construction details and statistics are provided in Appendix \ref{app:data}: we extract citation text paragraphs, limiting our paragraph selection such that the cited papers also belong to our reference data, and ensuring that full paper content and metadata are readily available for \emph{both} citing and cited papers. Using a set of rule-based heuristics we selected $5,971$ related work paragraphs -- comparable in size to the test set of \citet{Lu:2020}.

\textbf{Additional inputs.} We use this related work paragraph collection to extract \textbf{example sentences} for each cited paper. We created a pool consisting of citation sentences that cite the cited papers, drawn from all papers in the dataset. During experiments, we use this pool to select example sentences most similar to the gold reference paragraph via the SBERT model \cite{Reimers:2019}. Additionally, to steer generation, we enrich the reference paragraphs with \textbf{citation intents}. Intuitively, intents serve as a "hint" to reduce the possible space of generations and steer the LLM output towards the golden reference, inspired by expert recommendations for human authors \cite{Ridley:2012}. 

We experiment with two types of intents. \textbf{Categorical intents} were initially proposed to classify citation functions \cite{Jurgens:2018,Cohan:2019,Lauscher:2022} and assign categorical labels such as "Background" or "Motivation" to the citations. Although this type of intent is practical for classification tasks, the coarse schema inevitably leads to a loss of information that might be required to generate the actual related work paragraphs.\footnote{E.g., a coarse label "Background" covers free-form intents like \textit{``To compare the results of the proposed model with the results of the previous work''}, and \textit{``To provide a brief overview of the state of the art in argument mining''}, but fails to distinguish between the two.}

To address this, we experiment with novel machine-generated \textbf{free-form intents}, inspired by the numerous studies on how to write a literature review \cite{Pautasso:2013,Grant:2009,Randolph:2009}, the steps and principles of which can be applied to writing a related work section as well. \citet{Ridley:2012}  suggests using informal writing to prompt questions and to form the basis of the draft for the actual literature review, such as ``\emph{What are the methodological flaws of the previous methods?}''. Such informal writing corresponds to the free-form intents in our paper. To operationalize free-form intent, we define it as a sentence-level sequence briefly describing the reason why a particular paper is cited in a given paragraph. Figure \ref{fig:prompting} shows an example of a free-form intent. Note that both categorical and free-form intents are derived from the gold reference paragraph, and while free-form intents allow more flexibility, they also increase the risk of data leaks. We carry out extensive automatic and human analysis to make sure that our free-form intents do not have more n-gram overlap with gold references than abstracts of the original papers. Appendix \ref{app:intent} investigates this question in-depth and provides further examples of machine-generated free-form intents.

\subsection{Measurements}
We employ a range of measurements and metrics to characterize the citation texts generated by the LLMs in response to the prompt.

\textbf{Surface metrics.} To check whether model precisely follows formal requirements such as \textit{paragraph count} and \textit{citation mark} (e.g., \texttt{[REF\#1]}) given in the instructions, we report the average \textit{paragraph count} and percentage of utilization of \textit{citation mark} in generated citation texts. In addition, we report average word count and n-gram overlap between the input and the model output to check whether the model copies from the prompt.

\textbf{Conventional metrics.} To compare the generated text to the reference, we compute several conventional NLG metrics: ROUGE-L \cite{Lin:2004}, BERTScore \cite{Zhang:2020} and BLEURT \cite{Sellam:2020}. ROUGE is the most commonly used metric in prior work on citation text generation -- yet it only provides overlapping ratio between gold references and model outputs and lacks the capacity to evaluate the semantic correspondence between the two sequences. This is addressed by BERTScore and BLEURT metrics that use BERT-based \cite{Devlin:2019} representations to compare the generated text to the reference on the semantic level, showing greater robustness to paraphrases and better alignment with human assessments. We also compute BERTScore via SciBERT \cite{Beltagy:2019} to examine the effect of in-domain pre-training on the measurements.

\textbf{NLI-based metrics.} To measure the factual consistency between the gold reference and the output, we use two NLI models trained on curated fact-checking datasets. SummaC \cite{Laban:2022} generates NLI scores from the sentences of compared texts and calculates an overall score. TRUE makes binary decisions regarding entailment for a given textual pair\footnote{We use \{gold reference, model output\} as the NLI input.} \cite{Honovich:2022} and has been partially pre-trained on the SciTail \cite{Khot:2018} dataset in the scientific domain.

\section{Experiments}\label{sec:experiments}

We experiment with two state-of-the-art LLMs, open Llama 2-Chat 13B \cite{Touvron:2023-llama-2} and closed GPT 3.5 Turbo \cite{Ouyang:2022} (\texttt{gpt-3.5-turbo-0613-16k}). We dynamically construct our prompts according to the chosen input configuration. The outputs are analyzed using the measurement kit described above. To keep task complexity and computational costs at bay, we focus on the paragraphs discussing a \textbf{single cited paper}, resulting in 2,729 data instances. While using closed commercial LLMs is commonplace in NLP, this comes with reproducibility risks \cite{chen2023chatgpts}. To account for this, our Llama 2-Chat experiments were run on-site, taking approximately $\sim$30 hours on a single NVIDIA A100 GPU with 80GB memory, and are strictly reproducible. We used FlanT5-XXL\footnote{We used FlanT5-XXL to generate free-form intents for efficiency. Upon conducting a human evaluation, we determined that the quality of the generated output is satisfactory. More details can be found in Appendix \ref{app:intent}.} \cite{Chung:2022} to generate free-form intents for each citation paragraph in our dataset, and the fine-tuned MultiCite model\footnote{\url{https://huggingface.co/allenai/multicite-multilabel-scibert}}
\cite{Lauscher:2022} to assign categorical intents. Further details are given in Appendix \ref{app:impl}. We provide example model generations in Appendix \ref{app:model_gen_ex}.

\section{Results}\label{sec:results} 

\begin{figure*}
    \centering
    \includegraphics[width=0.96\textwidth]{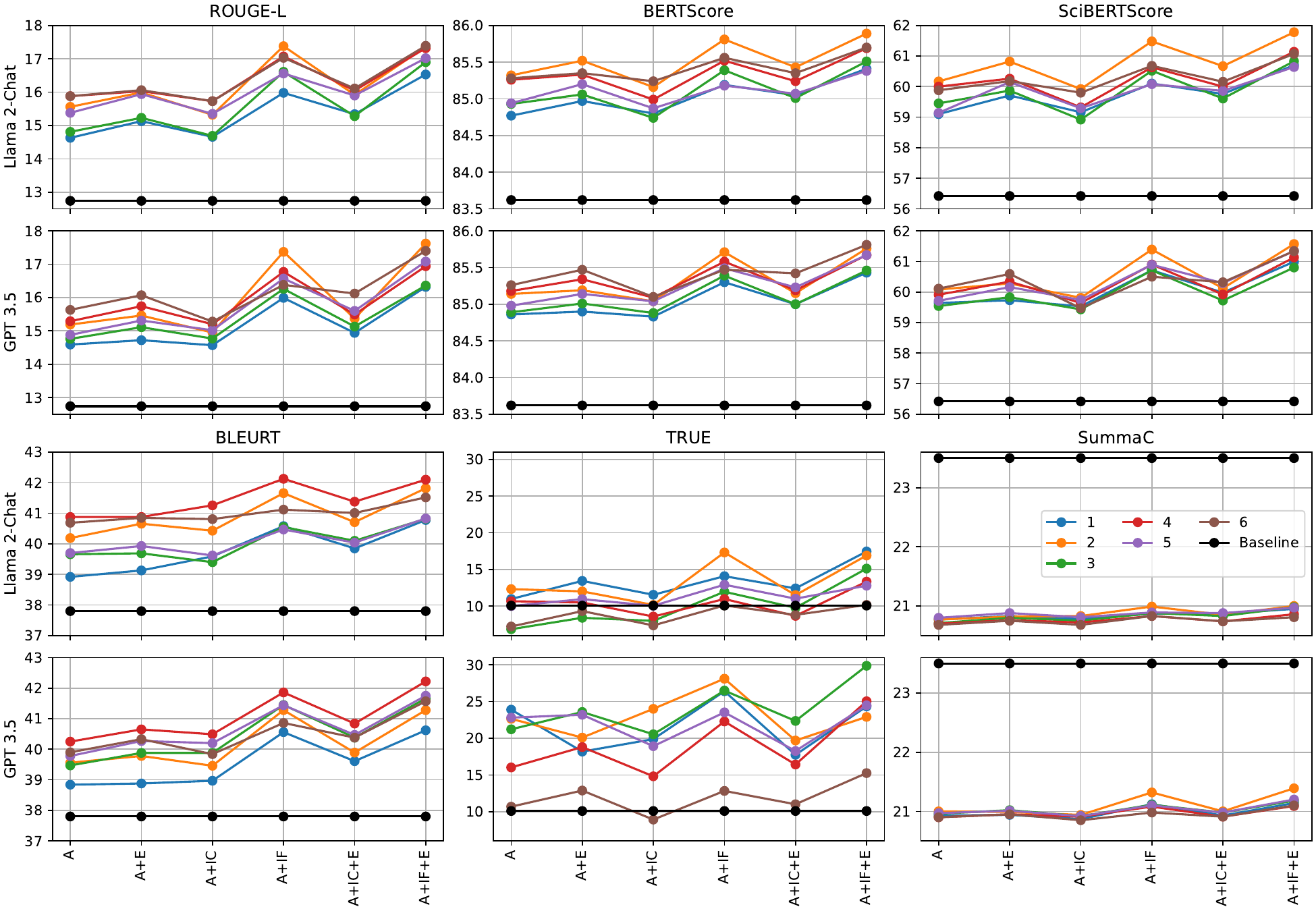}
    \caption{Conventional and NLI-based metric results. First two rows: ROGUE-L, BERTScore and SciBERTScore. Last two rows: BLEURT, TRUE and SummaC. Llama 2-Chat (13B) (above) and GPT 3.5 (below). \textbf{A}bstract + \textbf{I}ntent (\textbf{F}ree-form or \textbf{C}ategorical) + \textbf{E}xample, \#Instruction color-coded.}
    \label{fig:metric_results}
\end{figure*}

Our framework allows us to answer a range of questions about citation text generation in the context of modern LLMs. Here we focus on a subset of these questions and provide exhaustive experimental results in the Appendix \ref{app:impl}. We use the following notation to discuss experimental configurations: \texttt{\#(+A)(+IC/IF)(+E)}, where \texttt{\#} is the instruction template identifier ranging from $1$ to $6$ (Appendix \ref{app:prompts}), \texttt{+A} denotes source and target paper abstracts, \texttt{+IC} and \texttt{+IF} denote the categorical and free-form intents, and \texttt{+E} denotes an example citation sentence that cites the given paper. The instructions are adjusted to reflect the input components present in a given configuration. The example input in Figure \ref{fig:prompting} corresponds to the configuration \texttt{1+A+IF+E}. We concatenated both cited and citing abstracts and used it as baseline.

\textbf{RQ1: What is the impact of the input configuration on generated texts?} Figure \ref{fig:metric_results} presents our main results across different configurations. For all conventional metrics (ROGUE, (Sci)BERTScore, BLEURT), all configurations far outperform the baseline. For TRUE, while GPT 3.5 scores higher than baseline except for one configuration, Llama-2 performs close to the baseline. On the other hand, baseline SummaC scores are better than both model outputs. We also observe that providing the model with only abstracts (\texttt{+A}) yields the lowest degree of correspondence between the generated text and the reference in the vast majority of the configurations. This suggests that additional free-form intents and example sentences have a positive influence on performance in terms of both conventional and NLI-based measurements. We emphasize that this effect is consistent across different types of models and prompts. We also observe that providing models with free-form intents (\texttt{+IF}) increases the correspondence between generated and reference citation texts more than example sentences (\texttt{+E}). Jointly providing free-form intent and example (\texttt{+IF+E}) shows a combined effect and yields the best correspondence in 90.28\% of 72 measurements (two models $\times$ six prompts $\times$ six metrics). We additionally investigated whether this effect is only due to the input length and observed that even shorter \texttt{+A+IF+E} instances are still significantly ($p<0.05$) better than longer \texttt{+A} instances. Details of the experiments controlled for prompt length can be found in Appendix \ref{app:length_exp}. Overall, we note that the ranking of configurations remains mostly consistent across the task instructions and measurements. This suggests that the \emph{relative} performance of different input configurations might be estimated based on a small number of instruction variations.

\begin{table}
\small
\centering
\begin{tabular}{l|c|c|c|c}
\hline
\textbf{Model} & \textbf{NG-3} & \textbf{WC} & \textbf{PC} & \textbf{CM} \\
\hline
Llama 2-Chat & 25.29 & 125.95 & 1.31 & 67.64 \\
GPT 3.5 Turbo & 23.09 & 142.58 & 1.02 & 95.47 \\
\hline
Reference & - & 82.60 & 1.00 & 100.00 \\
\hline
\end{tabular}
\caption{Average surface measurements across the configurations \text{vs} single-paragraph single-citation reference data. NG-3: trigram overlap ratio (\%), WC: word count, PC: paragraph count, CM: citation mark usage (\%).}
 \label{tab:surf_results}
\end{table}

\textbf{RQ2: How do measurement scores differ between Llama 2 and GPT 3.5? } We see that the measurement type plays a significant role in revealing the difference between model outputs (Figure \ref{fig:metric_results}). Conventional metrics -- including the ROUGE score commonly used as the only evaluation metric by prior work -- fall short of distinguishing Llama 2-Chat and GPT 3.5 outputs. NLI-based metrics, however, do capture the difference: particularly for TRUE, the results are more pronounced in favor of GPT 3.5; a similar pattern can be observed for SummaC. As for the surface measurements, Table \ref{tab:surf_results} displays that GPT 3.5 tends to generate longer outputs than Llama 2-Chat; both models over-generate compared to the actual reference word count. We further observe that GPT 3.5 follows formal instructions more closely both for the paragraph count limitation and for consistent citation mark use.

\textbf{RQ3: How does the intent type affect the outputs?} Figure \ref{fig:metric_results} allows us to compare the performance of free-form and categorical intents across different experimental settings. We observe that categorical intents are insufficient to generate paragraphs that are close to the reference paragraphs. Categorical intent combinations with abstracts \texttt{(A+IC)} perform not better than only abstracts \texttt{(A)} in almost all cases. Furthermore, categorical intent combinations with example sentences \texttt{(A+IC+E)} are outperformed by using free-form intents \texttt{(A+IF)}, which holds for all measures and instructions, and for both investigated LLMs. This preliminary evidence suggests that while free-form intents are helpful, categorical intents might not be effective input components for citation text generation with LLMs.

\begin{figure}[t]
  \centering
  \includegraphics[width=1.05\columnwidth]{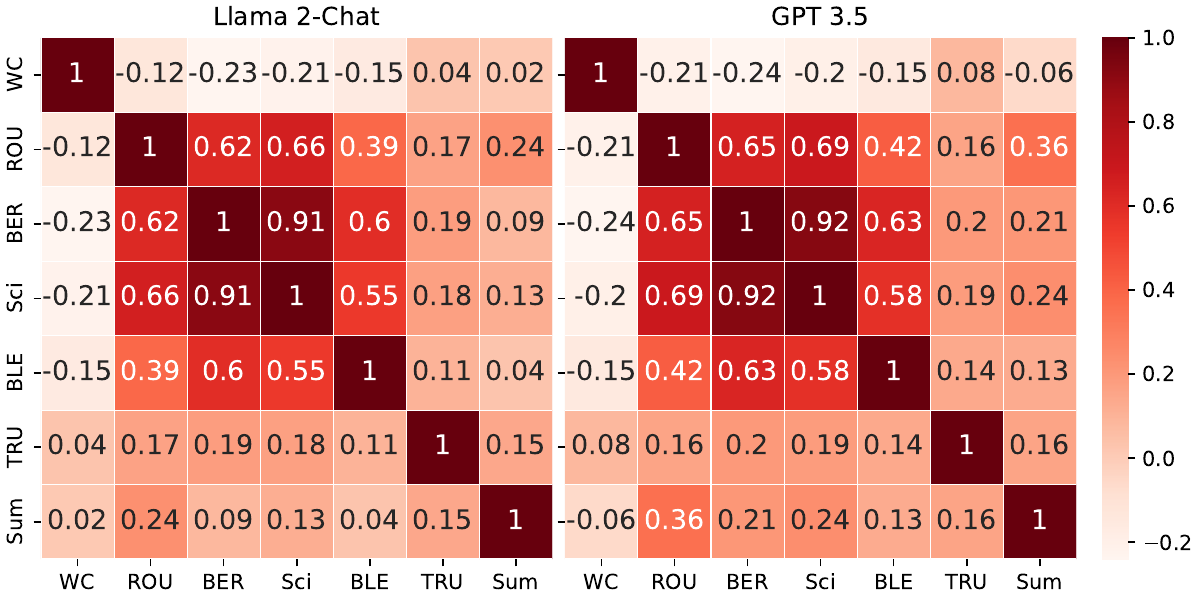}
  \caption{Pearson correlation between instance-level measurements over all configurations. WC: word count, Sum: SummaC.}
  \label{fig:correlations}
\end{figure}

\textbf{RQ4: What are the relationships between the measurements?} We observe that conventional metrics negatively correlate with the generated output length, while NLI metrics show little to no correlation (Figure \ref{fig:correlations}). This indicates that longer outputs per se do not result in having more similar meaning to gold reference and obtaining higher scores -- in line with the decreasing trend of ROUGE score after around 100 words observed by \citet{Sun:2019} and similar observations in citation text generation by \citet{Funkquist:2023}. We further see that conventional metrics show high correlations among themselves, but the correlations to the NLI-based metrics are low. TRUE and SummaC are less correlated with each other compared to conventional metrics. We hypothesize that since TRUE evaluates the entailment relation between two sequences in a binary manner, i.e. "entailment" or "contradiction", it might be sensitive to the changes in outputs. SummaC, on the other hand, processes paragraphs at the sentence level and produces an overall score by convolution -- decreasing its sensitivity but also leading to smaller differences between prompt configurations. These observations highlight the importance of multiple complementary measurements for citation text generation as opposed to the standard single-metric ROUGE-based evaluation.

\begin{table}
\small
\renewcommand{\arraystretch}{1.1}
\centering
\resizebox{\columnwidth}{!}{
\begin{tabular}{l|cccc|cc}
\hline
\textbf{Configuration} & \textbf{ROU} & \textbf{BER} & \textbf{Sci} & \textbf{BLE} & \textbf{TRU} & \textbf{Sum} \\
\hline
[H] 6+A & 14.16 & 85.69 & 61.20 & 37.36 & 10.00 & 21.33 \\
\hline
[H] 6+A+IF+E & \textbf{16.25} & \textbf{85.88} & \textbf{61.81} & \textbf{38.56} & \textbf{13.33} & \textbf{22.00} \\
\hline
\end{tabular}
}
\caption{Human generation experiment results.}
\label{tab:human_gen}
\end{table}

\section{Human Evaluation}\label{sec:humeval}

\begin{table}
\small
\renewcommand{\arraystretch}{1.1}
\centering
\begin{tabular}{l|c}
\hline
\textbf{Configuration} & \textbf{Coverage}\\
\hline
[LLama 2-Chat] 6+A & 0.40 \\
\hline
[LLama 2-Chat] 6+A+IF+E & 0.45 \\
\hline
[GPT 3.5 Turbo] 6+A & 0.35 \\
\hline
[GPT 3.5 Turbo] 6+A+IF+E & 0.49 \\
\hline
\end{tabular}
\caption{Human evaluation results: average fact coverage ratio on the sampled instances.}
\label{tab:human_eval}
\end{table}

\begin{table}
\small
\renewcommand{\arraystretch}{1.1}
\centering
\begin{tabular}{l|c|c}
\hline
\textbf{Metric} & \textbf{Correlation} & \textbf{p-value}\\
\hline
ROGUE & 0.103 & 0.265 \\
\hline
BERTScore & 0.253 & 0.005 \\
\hline
SciBERTScore & 0.232 & 0.010 \\
\hline
BLEURT & 0.289 & 0.001 \\
\hline
TRUE & 0.090 & 0.331 \\
\hline
SummaC & -0.046 & 0.621 \\
\hline
\end{tabular}
\caption{Pearson correlavation between human evaluation results and main metrics of the study.}
\label{tab:human_eval_corr}
\end{table}

\textbf{Human-generated citation texts.} To get further insights into the impact of input components on citation text generation, we conducted a small-scale human generation study. We sampled 30 instances from reference data used in our main experiments, for which three annotators with NLP backgrounds composed related work paragraphs. For each instance, annotators were first given the abstract information along with instruction \texttt{(6+A)} and wrote the citation paragraphs. Then, they repeated the same process with the addition of free-form intents and example sentences, corresponding to \texttt{(6+A+IF+E)}. The annotators had no access to gold reference paragraphs. The human-written texts were then measured using the metrics in our kit and compared against the reference. Table \ref{tab:human_gen} presents the results: we observe that the effect of including free-form intents and examples during citation text generation also holds when the texts are generated by human annotators. This suggests that the input components deemed necessary for the task can affect not only the LLM performance but also the outcomes of annotation and user studies.

\paragraph{Evaluation.} For the same set of instances, the respective Llama 2-Chat and GPT 3.5 outputs were manually evaluated by the annotators in terms of their correspondence to the gold reference. We used the pyramid method \cite{Nenkova:2004}, in which the generated output is compared to the gold reference in terms of its compliance with a set of basic facts derived from the reference. We extracted atomic facts from the reference paragraphs using GPT-3.5 (details in Appendix \ref{app:human}). The facts were manually quality-controlled and curated prior to evaluation. Then, for each citation paragraph instance, annotators checked whether the generated paragraphs mention the extracted atomic facts. The model and input configuration for the generated outputs were not known to the annotators, and the order of presentation of outputs was randomized for each question. Table \ref{tab:human_eval} demonstrates the average ratio of atomic fact coverage of each model. We observe that the benefit of including free-form intents and example sentences into the model input is supported by human evaluation. We note that human evaluation doesn't consistently favor any of the two LLMs, with the best results obtained by GPT 3.5 Turbo. Yet, we note that this might be due to the small scale of the human evaluation study, and leave further exploration of this discrepancy to the future. Further details of the both human generation and evaluation can be found in Appendix \ref{app:human}. We additionally investigated the correlation between human evaluation results and other metrics. Table \ref{tab:human_eval_corr} shows that ROGUE, SummaC and TRUE have no significant correlation with human scores. Although (Sci)BERTScore and BLEURT have statistically significant correlations, they are not high. This suggests that human evaluation in terms of atomic facts yields a complementary perspective that needs to be considered while evaluating generated citation texts.

\paragraph{Qualitative observations.} Our evaluations yielded few informal insights which we deem useful for follow-up research. 
LLM generations were typically more verbose (see Table \ref{tab:surf_results}), but also less specific. We observed that the wording of the instruction affects the \emph{pragmatics} of the generated paragraph: for some instructions, the model tended to generate a text \textit{comparing} two papers (\textit{"While the main paper does X, the related paper does Y"}), instead of \textit{discussing} one paper in the context of the other (\textit{"Unlike our paper, [REF\#1] has done Y"}). As this is not reflected in the metric performance scores, we hypothesize that pragmatic mismatch might not be captured by the automatic evaluation metrics. We found that the success of generations depended on the content of the gold reference: while high-level discussion of related work can be generated from the abstracts, going into specifics of a paper requires information not available in the input. Uninformative abstracts were also hard to generate from, both for humans (who wrote short and uninformative citation texts in response) and for LLMs (that were forced to hallucinate text). Since the setting of our human study is insufficient to investigate these observations empirically, we leave this exploration for future research.

\section{Conclusion} 

The last generation of LLMs has enabled a wide range of novel, creative NLG tasks characterized by flexible input component space and a wide range of plausible outputs. This flexibility brings about new challenges as it introduces additional degrees of freedom into the experimental setup, but it also allows NLP researchers to systematically study the relationships between the input components, the instructions, and the generated outputs. In this work, we have addressed the variability of creative NLG tasks by systematically exploring the task of citation text generation. We proposed a framework for systematically comparing task configurations and used it to study the impact of task input and instruction on the citation text generation performance, by both LLMs and humans, using a wide-spanning range of measurements to characterize the LLM outputs. Our insights contribute to a better understanding of the role of input configurations, instructions, and output measurements in LLM-based language processing, and our framework facilitates the study of citation text generation in the age of LLMs. Our human evaluations provide additional insights to guide future work. In an environment where anecdotal evidence of LLMs' impressive capabilities is overabundant, our work contributes to the best practice for the systematic study of LLM-based approaches to complex, creative text-generation tasks.

\section{Limitations} \label{sec:limitations}

\textbf{Language and domain.} Our experiments are limited to the English papers from the ACL Anthology. English is the standard language of communication in most research fields, and the focus on English is typical for scholarly NLP. ACL Anthology was chosen due to its availability, open licensing, and familiarity of the research area to the paper authors. Applying our approach in a cross-lingual and multi-lingual setting and novel domains is an engaging future work direction that can be pursued once the necessary research infrastructure is available.

\paragraph{Free-form intents and lexical overlap.} In Section \ref{sec:results}, we compare free-form and categorical intents. Manually creating a citation intent for each dataset instance is not feasible because intents should be specific to each citation, not in a generalized and abstract form. Since it is a laborious task for humans, we generate free-form intents from the gold reference paragraphs using a Flan-T5 model and predict categorical intents using an off-the-shelf model from prior work. This raises a concern that the predicted intents might leak information from the gold reference paragraph, especially pronounced for free-form intents. To investigate the extent to which this is the case, we conducted extensive additional experiments on free-form intents (Appendix \ref{app:intent}). We observed that the lexical overlap of free-form intents is in fact lower than the lexical overlap already present in the abstracts. The subsequent small-scale manual analysis confirmed that the generated intents do not contain sufficient information to re-create the reference paragraph (Appendix \ref{app:intent}). Exploring ways to further increase the informativeness of citation intents (categorical or free-form), while minimizing information leaks is an open avenue for follow-up research.

\paragraph{Competing with state of the art.} We explicitly do not compare to prior systems, since the goal of our work is to study the \textit{effect} of the input configuration and instructions, and \textit{not} to produce a top-performing model instance. Given the capabilities of modern LLMs, a side-by-side comparison could put earlier systems at a disadvantage and would conflate a wide range of potential sources of improvement \cite{Nityasya:2023}. Following our results, arriving at a best-performing citation text generation system would need further extensive experimentation with instructions (Section \ref{sec:related_work}). Furthermore, although  an overall score is beneficial for benchmarking purposes, defining such score would overshadow the importance of investigating individual components of the complex tasks.

\paragraph{Human evaluation.} We stress that our human evaluation study is exploratory, and its results would require large-scale validation on a wider annotator base. Our results and qualitative insights can serve as a basis to build better citation text generation models in the future. 

\paragraph{Limitations of the setup.} To keep our study tractable, we had to impose limitations on our setup. We limited our experiments to paragraphs containing a single citation since paragraphs containing multiple citations would require substantially longer inputs to include all information necessary for generation. This also prevents us from utilizing in-context-learning framework. This limitation can be revisited once more computationally efficient open LLMs with higher input lengths become available. Due to the computational costs of LLM experimentation, including more LLMs into analysis would mean compromising on the rigor of the experimental setup.
While we put effort into validating our findings using a range of instructions instead of a single prompt, adding more instructions would allow us to further verify our findings and get better estimates of the absolute performance. We thus recommend expanding the instruction pool for the follow-up work that aims to produce a best-performing system. In our experiments, we considered three groups of input components: abstracts, intents, and example sentences. This set can be easily extended based on the reference data released along with this paper, which contains \emph{rich metadata} and pointers to \emph{parsed full papers for both citing and cited works}, with \emph{one and multiple citations per paragraph}.

\section*{Ethics Statement}

We believe that systematically studying the relationship between the input components, instructions and LLM outputs for creative NLG tasks is crucial for the understanding of how LLMs work and what factors influence their behavior. The data and models used in this study -- apart from GPT 3.5 Turbo -- are publicly available and distributed under open licenses, facilitating long-term reproducibility and allowing the community to build upon our work. No external human annotators were involved in the study. The task of citation text generation aims to increase the efficiency of scientific work. While misuse of citation text generation could cause reduced engagement with scientific literature, we believe that using such systems as an aid -- not a replacement for paper reading -- can facilitate exploration of vast scientific literature, and that the benefits of such systems would outweigh the risks. 

\section*{Acknowledgements}

This work has been funded by the LOEWE Distinguished Chair “Ubiquitous Knowledge Processing”, LOEWE initiative, Hesse, Germany (Grant Number: LOEWE/4a//519/05/00.002(0002)/81), and by the German Research Foundation (DFG) as part of the PEER project (grant GU 798/28-1), and by the European Union (ERC, InterText, 101054961). Views and opinions expressed are however those of the author(s) only and do not necessarily reflect those of the European Union or the European Research Council. Neither the European Union nor the granting authority can be held responsible for them.  We gratefully acknowledge the support of Microsoft with a grant for access to OpenAI GPT models via the Azure cloud (Accelerate Foundation Model Academic Research). Yufang Hou is supported by the Visiting Female Professor Programme from Technical University of Darmstadt.

\bibliography{custom}
\bibliographystyle{acl_natbib}

\appendix

\section{Dataset}
\label{app:data}

\subsection{Title List}\label{app:title}

List of related work titles used in dataset creation is as follows. 

\{\textit{"related work", "related works", "previous work", "background", "introduction and related works", "introduction and related work", "background and related work", "background and related works", "previous related work", "previous related works", "backgrounds", "previous and related work", "previous and related works"}\}

\subsection{Cleaning and Post-processing} \label{app:clean}

We performed several additional cleanup operations on the instances extracted from parsed ACL Anthology dataset. We removed instances with corrupted components e.g., abstract, metadata, citation mark, paragraphs. We encountered papers that were published in different venues with the same title and abstract. Such duplicates were removed. A small number of non-English papers were removed. We used word count threshold of 40 for extracted paragraphs and of 10 for citation sentences to filter out erroneous citation paragraphs due to PDF parsing issues. The cleanup was applied to the related work paragraph dataset and to the example citation sentence dataset in parallel. If there were no instances left for a cited paper after the cleanup, citation sentences for that paper were also removed from the example sentence pool.

Some cited paper's citation sentences are not included in the example sentence dataset due to our cleanup procedure. For instance, corresponding sentences may not be segmented well or their length may be below the token threshold. To extract sentences from the paragraphs, we used the \textit{scispacy}\footnote{\url{https://allenai.github.io/scispacy/}}. While determining the most similar example citation sentence, we used the \textit{all-MiniLM-L6-v2} sentence transformer model\footnote{\url{https://huggingface.co/sentence-transformers/all-MiniLM-L6-v2}}.

\subsection{Dataset Staticstics}\label{app:stats}

Tables \ref{tab:rw_stats} and \ref{tab:ex_stats} show core statistics for the resulting self-contained collection of related work paragraphs along with the respective papers that they cite and example citation sentences.

\begin{table}[h]
\centering
\small
\renewcommand{\arraystretch}{1.0}
\begin{tabular}{l|r}
\hline
Paragraphs & 5,971 \\
Total citation & 12,950 \\
Unique citing papers & 4,605 \\
Unique cited papers & 6,620 \\
Avg. occur. of a cited paper & 1.96 \\
Sentence count per paragraph & 4.22 \\
Word count per paragraph & 98.67 \\ 
\hline
\end{tabular}
\caption{Related work paragraph dataset statistics}
 \label{tab:rw_stats}
\end{table}

\begin{table}[h]
\centering
\small
\renewcommand{\arraystretch}{1.0}
\begin{tabular}{l|r}
\hline
Sentences & 73,139 \\
Unique citing papers & 16,338 \\
Unique cited papers & 6,594 \\
Sentence per cited paper & 11.05 \\
Word count per sentence & 35.30 \\ 
\hline
\end{tabular}
\caption{Example sentence dataset statistics}
 \label{tab:ex_stats}
\end{table}

\begin{figure}[ht]
  \centering
  \includegraphics[width=\columnwidth]{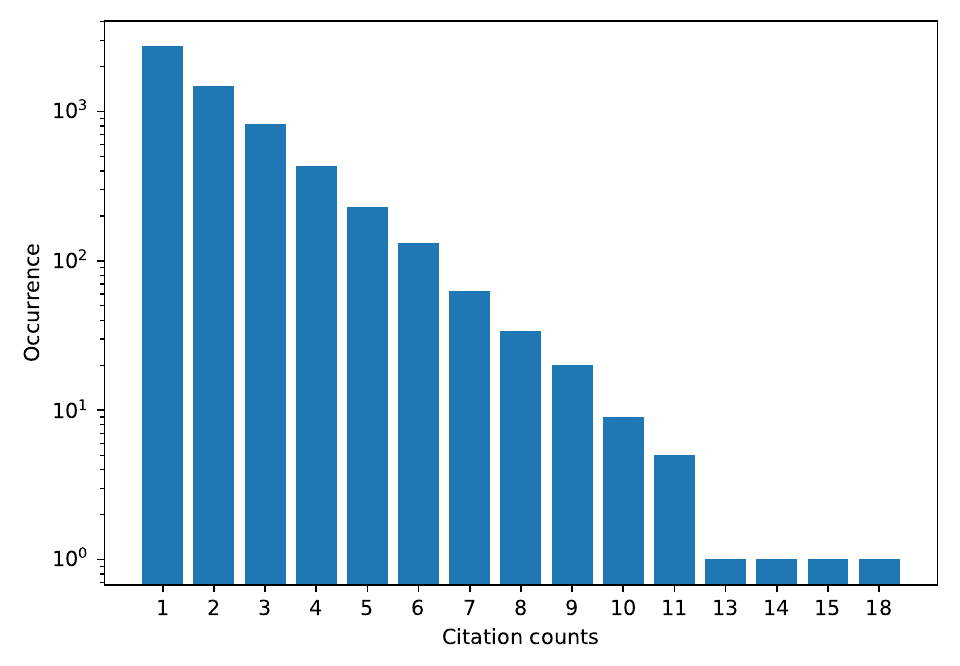}
  \caption{Citation count distribution in logarithmic scale}
  \label{fig:citation_count}
\end{figure}

The distribution of the citation counts in the paragraphs is shown in Figure \ref{fig:citation_count}. 
Around 2,700 paragraphs include only one citation and the most crowded paragraphs include up to 18 citations. In this work, we focus on the subset of paragraphs that include only one citation.

\subsection{Dataset fields}\label{app:columns}

Field names along with their descriptions for the related work paragraph and the citation sentence datasets are given in Tables \ref{tab:column_rw} and \ref{tab:column_ex}, respectively.

\begin{table*}
\centering
\small
\renewcommand{\arraystretch}{1.2}
\begin{tabular}{l|p{0.7\textwidth}}
\hline
\textbf{Column name} & \textbf{Description} \\
\hline
\verb|acl_id| & Unique ACL ID of the citing paper. Since a paper can have different related work paragraphs that satisfy conditions, there can be instances with the same \verb|acl_id|. Although it is a unique identifier for distinguishing papers in ACL Anthology, this is not a unique identifier for this dataset. This rule is also valid for other citing paper meta features. \\
\verb|abstract| & Abstract of the citing paper. \\
\verb|corpus_paper_id| & Semantic Scholar ID of the citing paper. \\
\verb|pdf_hash| & sha1 hash of the PDF. \\
\verb|numcitedby| & The citing paper’s citation count based on Semantic Scholar. \\
\verb|url| & URL of the citing paper. \\
\verb|publisher| & Publisher of the citing paper. \\
\verb|address| & Address of the conference or venue. \\
\verb|year| & The citing paper’s publication year. \\
\verb|month| & The citing paper’s publication month. \\
\verb|booktitle| & The name of the proceedings if it is a conference paper. \\
\verb|author| & Authors of the citing paper. \\
\verb|title| & Title of the citing paper. \\
\verb|pages| & Page information of citing paper. \\
\verb|doi| & DOI identifier of the citing paper. \\
\verb|number| & Article number of the citing paper if it is a journal paper. \\
\verb|volume| & Volume number of the citing paper if it is a journal paper. \\
\verb|journal| & Journal name of the citing paper if it is a journal paper. \\
\verb|editor| & Name of the editors if it is a journal paper. \\
\verb|isbn| & ISBN number of the citing paper. \\
\verb|paragraph_xml| & Citation paragraph with XML tags. It also includes other information about the citations relative to citing paper. \\
\verb|paragraph| & Citation paragraph without XML tags. Like normal text in an article. \\
\verb|cited_paper_marks| & This includes XML tags of target cited papers relative to citing papers. Identifiers are not absolute but relative. These tags also exist in \verb|paragraph_xml| column. Since there can be multiple cited papers in the paragraph each mark is separated by " \verb|%%%| " (space + 3 consecutive \verb|%| + another space) . \\
\verb|cited_paper_titles| & Titles of the cited papers separated by " \verb|%%%| ". \\
\verb|cited_papers_acl_ids| & \verb|acl_ids| of the cited papers separated by " \verb|%%%| ". \\
\verb|cited_papers_abstracts| & Abstracts of the cited papers separated by " \verb|%%%| ". \\
\hline
\end{tabular}
\caption{Column names and descriptions for the related work paragraph dataset.}
\label{tab:column_rw}
\end{table*}

\begin{table*}
\small
\centering
\renewcommand{\arraystretch}{1.2}
\begin{tabular}{l|p{0.8\textwidth}}
\hline
\textbf{Column name} & \textbf{Description} \\
\hline
\verb|example_id| & Unique id of the example sentence instances. Its construction formula is \verb|acl_id| of cited paper + "\verb|%|" + extraction order number. \\
\verb|sentence| & Example sentence citing target cited paper. \\
\verb|paragraph_xml| & XML version of the paragraph which example sentence belongs to. (From the related work section of the citing paper) \\
\verb|paragraph| & Textual version of the paragraph which example sentence belongs to. (From the related work section of the citing paper) \\
\verb|citation_mark| & This includes XML tags of target cited paper’s citation marks. \\
\hline
\end{tabular}
\caption{Column names and descriptions for the example citation sentence dataset. The dataset also includes metadata of the citing and the cited papers as given in Table \ref{tab:column_rw}.}
 \label{tab:column_ex}
\end{table*}

\subsection{Intent Generation}\label{app:intent}

\begin{table}[htbp!]
\small
\centering
\begin{tabular}{c|c|c}
\textbf{N-gram} & \textbf{Gold vs. Intent} & \textbf{Gold vs. Abstract} \\ \hline
1 & 0.10 & 0.25 \\ \hline
2 & 0.06 & 0.05 \\ \hline
3 & 0.04 & 0.02 \\ 
\end{tabular}
\caption{Average ratio of the number of overlapping n-grams of gold paragraphs with intents (abstracts) to the total number of gold paragraph n-grams, stop words excluded.}
\label{tab:intent_ngram}
\end{table}

For intent generation we experimented with a range of models such as LLaMA (7B) \cite{Touvron:2023-llama}, Alpaca (7B) \cite{Taori:2023} and BLOOMZ (7.1B) \cite{Muennighoff:2023}. Yet, models other than \textit{FlanT5} did not yield meaningful outputs, e.g. occasionally generating random character sequences.

We conducted preliminary experiments for intent generation on a subsample of our dataset, exploring both zero-shot and few-shot configurations. In the zero-shot setting, we instructed the models to generate intent of the given target paragraph without showing any examples. In few-shot setting, we provided two-three paragraphs and their corresponding intents. To generate example paragraph-intent pairs, we conducted 100 zero-shot generations and manually selected six examples that successfully reflect the intent of the paragraph. We observed that in the few-shot setting the models tended to copy the examples into the output. Therefore, we settled on a \textit{zero-shot setting} as our final configuration to generate the intent. We use the following FlanT5 prompt:

\noindent \textit{What is intention of the following paragraph? \\
\{Target paragraph\}}   

We investigated several decoding strategies to optimize generations such as greedy search, beam search, multinomial sampling, multinomial sampling with beam search and contrastive search with different hyperparameters. In the final setting, we opted for \textit{greedy decoding} due to its output quality and reproducibility of the outputs. 

To examine whether generated intents copy from the gold paragraph, we conducted n-gram analysis. We calculated the ratio of n-gram overlap between intents and gold references, and compared it to the overlap between abstracts and gold references. As demonstrated in Table \ref{tab:intent_ngram}, abstracts -- a universally used input type in citation text generation -- have a higher overlap with the gold paragraphs than generated intents do, and for the bigrams and trigrams the results are head-to-head. This implies that our unstructured intents do not reveal substantially more keyword information than the already-present abstracts. To follow up, we randomly sampled 100 intents and conducted a human evaluation to determine whether an intent discloses important information of the gold reference. Out of 100 instances observed, only in two cases intents reveal more information than expected. For example,

Paragraph: \noindent \textit{There is a long tradition of work on the within document coreference (WDC) problem in NLP, which links named entities with the same referent within a document into a WDC chain. State-of-the-art WDC systems, e.g. (Ng and Cardie, 2001), leverage rich lexical features and use supervised and unsupervised machine learning methods.}

Intent: \textit{This paper proposes a novel approach to WDC that leverages the richness of the document and the richness of the lexicon.}

Although there are overlapping words between paragraph and intent, the intent is not sufficient to reconstruct original paragraph. However, the vast majority of intents used in this work are not as specific as in those very rare cases. Below is a random sample of machine-generated intents used in our study:

\begin{itemize}
[leftmargin=*,noitemsep,nolistsep]
\item To describe the state of the art in WSD systems.
\item To describe the Universal Dependency project.
\item To provide a comparison of the pruning distances for dependency-based relation extraction models.
\item To describe the work
\item To describe the problem and the solution.
\item To describe the crowdsourcing approach used to bootstrap YARN.
\item To describe the relation between Nominal SRL and SemEval.
\item To provide a brief overview of the state-of-the-art in unsupervised structured prediction.
\item To compare the performance of our approach with Yarowsky et al. (2001) and other related work.
\item To introduce naive, linguistically motivated regularization methods such as sentence length, punctuation and word frequency.
\item To provide a comparison of UDon2 and Udapi.
\item To present a new technique for combining NMT models that is capable of addressing i and ii.
\item To describe the work
\item To describe a study.
\item To provide a brief overview of the state of the art in multilingual representation learning.
\item To describe the problem of query expansion
\item To provide a brief review of the related works.
\item To describe the state of the art in multilingual model evaluation.
\item To describe an email thread summarization approach.
\end{itemize}

\section{Task instruction templates} \label{app:prompts}

Models that we use in related work paragraph generation take prompts in two segments: \emph{system prompt} and \emph{user message}. System prompt is a fixed instruction for each session to guide the model how to react to user messages. User message contains additional information related to the instance at hand. 
In most cases we use system prompt to provide the task instruction, and use the user message to provide instance-specific data -- Template 2 is an exception in that  there input components are embedded into the user message, and system prompt remains empty. To increase the diversity of the templates, we used different prompting strategies such as direct instruction, chain-of-thought, role-playing, and instruction list. The following subsections exemplify the system inputs used in our work for the case where all input components are included into the instruction. 

\subsection{Template 1 (Direct Instruction)}

\begin{itemize}
    \item[] \textbf{System prompt:} \textit{Your aim is to generate an exactly single paragraph to be used in related work section in a main paper. You will be given the main paper's abstract and a relevant paper's abstract. The paragraph should reflect the intent and you need to refer the relevant paper in the same paragraph by using citation mark [REF\#1]. You can inspire from the given example.}
    \item[] \textbf{Custom instance prompt:} \textit{Main paper abstract: \{Citing paper abstract\} \\
    Relevant paper abstract: \{Cited paper abstract\} \\
    Intent: \{Intent of the paragraph\} \\
    Example: \{Example citation sentence\}}
\end{itemize}

\subsection{Template 2 (Chain-of-thought)}

\begin{itemize}
    \item[] \textbf{System prompt:} - 
    \item[] \textbf{Custom instance prompt:} \textit{Assume that you are the author of a paper whose abstract is as follows: \\
    \{Citing paper abstract\} \\
    In your paper's related work paragraph, you want to cite a paper whose abstract is as follows: \\
    \{Cited paper abstract\} \\
    Intent of the related work paragraph should be as follows: \\
    \{Intent of the paragraph\} \\
    You can inspire from the given example: \\
    \{Example citation sentence\} \\
    How would you write an exactly one related work paragraph for this purpose? While citing use the citation mark [REF\#1]. Your output must strictly consist of the related work paragraph only, nothing else.}
    
\end{itemize}

\subsection{Template 3 (Instruction List)}

\begin{itemize}
    \item[] \textbf{System prompt:} \textit{Follow given instructions:
        1-) You will be given main paper's abstract, a relevant paper's abstract, an intent and an example sentence. \\
        2-) Write a related work paragraph that is belonging to main paper and citing relevant paper. \\
        3-) The goal of your paragraph should be the given intent. \\
        4-) You can utilize example sentence as how the relevant paper is cited before. \\ 
        5-) Start your paragraph without any other explanations.\\
        6-) Use [REF\#1] as citation mark. \\
        7-) Your output should consist of exactly single paragraph.}
    \item[] \textbf{Custom instance prompt:} \textit{Main paper abstract: \{Citing paper abstract\} \\
    Relevant paper abstract: \{Cited paper abstract\} \\
    Intent: \{Intent of the paragraph\} \\
    Example: \{Example citation sentence\}}
    
\end{itemize}

\subsection{Template 4 (Role-playing)}

\begin{itemize}
    \item[] \textbf{System prompt:} \textit{You are writing a research paper and want to discuss another, related paper, with a certain intent -- the purpose of the discussion. Generate exactly one paragraph of text that discusses the related paper in context of the main paper and follows the intent. You will be given the main paper abstract, the related paper's abstract, and the intent sentence. You can also utilize the given example sentence. Refer to the related paper by using a citation mark [REF\#1]. You should generate exactly one paragraph of text, nothing else.}
    \item[] \textbf{Custom instance prompt:} \textit{Main paper abstract: \{Citing paper abstract\} \\
    Relevant paper abstract: \{Cited paper abstract\} \\
    Intent: \{Intent of the paragraph\} \\
    Example: \{Example citation sentence\}}
    
\end{itemize}

\subsection{Template 5 (Role-playing)}

\begin{itemize}
    \item[] \textbf{System prompt:} \textit{Imagine that you are a scientist writing a research paper. Your goal is to write a related work paragraph that discusses the related paper in context of your main paper. The related paper should be mentioned in the paragraph by using a citation mark [REF\#1]. You will be given the main paper abstract, the related paper abstract, and the intent -- the reason why you are citing the paper. An example sentence is also given to show how the related paper has been cited before. Your output should consist of exactly one paragraph of text and include the citation mark.}
    \item[] \textbf{Custom instance prompt:} \textit{Main paper abstract: \{Citing paper abstract\} \\
    Relevant paper abstract: \{Cited paper abstract\} \\
    Intent: \{Intent of the paragraph\} \\
    Example: \{Example citation sentence\}}
    
\end{itemize}

\subsection{Template 6 (Direct Instruction)}

\begin{itemize}
    \item[] \textbf{System prompt:} \textit{You are given two research papers: main paper and related paper. Generate one paragraph of text that discusses the related paper in the context of the main paper, given the intent -- the reason why the main paper discusses the related paper. A citation sentence is also given to be taken as example. Use a citation mark [REF\#1] to refer to the related paper. Your output should consist of exactly one paragraph of text and include the citation mark.}
    \item[] \textbf{Custom instance prompt:} \textit{Main paper abstract: \{Citing paper abstract\} \\
    Relevant paper abstract: \{Cited paper abstract\} \\
    Intent: \{Intent of the paragraph\} \\
    Example: \{Example citation sentence\}}
    
\end{itemize}

\begin{table*}[ht]
\centering
\renewcommand{\arraystretch}{1.1}
\resizebox{0.95\textwidth}{!}{
\begin{tabular}{l|c|c|c|c|c|c!{\vrule width 2pt}c|c|c|c!{\vrule width 2pt}c|c}
\hline
\multirow{2}{*}{\textbf{Configuration}} & \multicolumn{6}{c!{\vrule width 2pt}}{\textbf{Surface}} & \multicolumn{4}{c!{\vrule width 2pt}}{\textbf{Conventional}} & \multicolumn{2}{c}{\textbf{NLI}} \\
\cline{2-13}
& \textbf{NG-1} & \textbf{NG-2} & \textbf{NG-3} & \textbf{WC} & \textbf{PC} & \textbf{CM} & \textbf{ROUGE-L} & \textbf{BERTScore} & \textbf{SciBERTScore} & \textbf{BLEURT} & \textbf{TRUE} & \textbf{SummaC} \\
\hline
Abs. Baseline & - & - & - & 244.20 & - & - & 12.74 & 83.62 & 56.42 & 37.80 & 10.08 & 23.50 \\
\hline
1+A & 61.48 & 37.38 & 26.70 & 139.88 & 1.50 & 30.69 & 14.63 & 84.77 & 59.10 & 38.92 & 10.98 & 20.77 \\
1+A+E & 63.07 & 37.94 & 26.97 & 139.45 & 1.64 & 74.36 & 15.13 & 84.97 & 59.71 & 39.13 & 13.45 & 20.83 \\
\rowcolor{lightgray} 1+A+IC & 59.04 & 34.13 & 23.76 & 136.22 & 1.57 & 45.30 & 14.66 & 84.80 & 59.16 & 39.59 & 11.56 & 20.78 \\
1+A+IF & 59.81 & 34.88 & 24.09 & 137.63 & 1.48 & 41.62 & 15.98 & 85.19 & 60.10 & 40.58 & 14.09 & 20.87 \\
\rowcolor{lightgray} 1+A+IC+E & 60.82 & 35.09 & 24.45 & 135.74 & 1.63 & 80.15 & 15.34 & 85.05 & 59.76 & 39.85 & 12.43 & 20.87 \\
1+A+IF+E & 61.29 & 35.57 & 24.56 & 137.23 & 1.63 & 77.54 & 16.53 & 85.41 & 60.70 & 40.78 & 17.46 & 20.95 \\
\hline
2+A & 64.78 & 37.02 & 26.04 & 110.64 & 1.08 & 63.07 & 15.56 & 85.32 & 60.17 & 40.19 & 12.33 & 20.77 \\
2+A+E & 64.52 & 34.94 & 23.74 & 107.81 & 1.11 & 82.87 & 15.99 & 85.52 & 60.82 & 40.67 & 12.02 & 20.82 \\
\rowcolor{lightgray} 2+A+IC & 62.97 & 35.05 & 24.41 & 115.98 & 1.30 & 86.75 & 15.32 & 85.16 & 59.92 & 40.43 & 10.17 & 20.83 \\
2+A+IF & 64.94 & 37.21 & 26.11 & 115.08 & 1.11 & 91.30 & 17.38 & 85.81 & 61.48 & 41.66 & 17.33 & 20.99 \\
\rowcolor{lightgray} 2+A+IC+E & 64.27 & 35.08 & 24.06 & 111.01 & 1.18 & 88.57 & 15.92 & 85.43 & 60.67 & 40.71 & 11.51 & 20.85 \\
2+A+IF+E & 64.79 & 35.89 & 24.52 & 113.89 & 1.15 & 89.71 & 17.36 & 85.89 & 61.78 & 41.82 & 16.91 & 21.00 \\
\hline
3+A & 61.52 & 36.09 & 25.37 & 121.42 & 1.31 & 37.56 & 14.81 & 84.93 & 59.45 & 39.66 & 6.86 & 20.71 \\
3+A+E & 64.01 & 38.30 & 27.33 & 125.22 & 1.48 & 76.25 & 15.23 & 85.06 & 59.87 & 39.69 & 8.42 & 20.80 \\
\rowcolor{lightgray} 3+A+IC & 57.62 & 31.62 & 21.55 & 124.33 & 1.41 & 28.16 & 14.69 & 84.74 & 58.92 & 39.40 & 8.00 & 20.75 \\
3+A+IF & 61.24 & 36.29 & 25.54 & 125.82 & 1.32 & 28.42 & 16.61 & 85.39 & 60.52 & 40.55 & 11.97 & 20.88 \\
\rowcolor{lightgray} 3+A+IC+E & 61.86 & 35.31 & 24.69 & 126.10 & 1.51 & 76.90 & 15.28 & 85.01 & 59.61 & 40.10 & 9.78 & 20.83 \\
3+A+IF+E & 63.28 & 37.91 & 26.93 & 130.84 & 1.47 & 75.52 & 16.90 & 85.51 & 60.83 & 40.82 & 15.13 & 20.98 \\
\hline
4+A & 62.03 & 35.18 & 24.30 & 121.93 & 1.01 & 54.55 & 15.88 & 85.26 & 60.00 & 40.88 & 10.68 & 20.70 \\
4+A+E & 64.61 & 37.85 & 26.61 & 124.53 & 1.03 & 82.07 & 16.03 & 85.33 & 60.26 & 40.88 & 10.51 & 20.76 \\
\rowcolor{lightgray} 4+A+IC & 59.50 & 31.94 & 21.69 & 125.68 & 1.00 & 40.49 & 15.73 & 84.99 & 59.32 & 41.26 & 8.59 & 20.72 \\
4+A+IF & 61.58 & 35.36 & 24.35 & 128.27 & 1.02 & 42.73 & 17.07 & 85.52 & 60.63 & 42.13 & 10.98 & 20.83 \\
\rowcolor{lightgray} 4+A+IC+E & 61.70 & 33.38 & 22.57 & 125.64 & 1.03 & 81.43 & 16.05 & 85.24 & 60.00 & 41.38 & 8.68 & 20.74 \\
4+A+IF+E & 63.31 & 36.48 & 25.18 & 129.24 & 1.05 & 78.56 & 17.32 & 85.69 & 61.14 & 42.10 & 13.35 & 20.86 \\
\hline
5+A & 63.41 & 40.21 & 30.04 & 128.45 & 1.40 & 25.95 & 15.38 & 84.94 & 59.41 & 39.70 & 10.02 & 20.80 \\
5+A+E & 63.85 & 38.96 & 28.42 & 123.76 & 1.58 & 76.99 & 15.94 & 85.20 & 60.15 & 39.93 & 10.96 & 20.88 \\
\rowcolor{lightgray} 5+A+IC & 60.97 & 36.90 & 26.75 & 130.93 & 1.56 & 36.64 & 15.35 & 84.87 & 59.29 & 39.62 & 10.06 & 20.81 \\
5+A+IF & 61.17 & 37.29 & 27.02 & 132.86 & 1.56 & 30.74 & 16.56 & 85.18 & 60.08 & 40.47 & 12.93 & 20.89 \\
\rowcolor{lightgray} 5+A+IC+E & 61.45 & 36.19 & 25.97 & 129.90 & 1.76 & 78.19 & 15.90 & 85.07 & 59.86 & 40.04 & 11.03 & 20.88 \\
5+A+IF+E & 61.73 & 36.92 & 26.45 & 130.96 & 1.77 & 76.20 & 17.02 & 85.38 & 60.64 & 40.83 & 12.80 & 20.97 \\
\hline
6+A & 62.98 & 34.84 & 23.55 & 122.18 & 1.01 & 92.55 & 15.88 & 85.28 & 59.88 & 40.69 & 7.23 & 20.68 \\
6+A+E & 64.79 & 36.37 & 24.88 & 125.88 & 1.07 & 95.34 & 16.06 & 85.35 & 60.18 & 40.85 & 9.33 & 20.75 \\
\rowcolor{lightgray} 6+A+IC & 62.09 & 34.02 & 23.02 & 123.94 & 1.02 & 89.94 & 15.73 & 85.24 & 59.80 & 40.81 & 7.38 & 20.68 \\
6+A+IF & 64.60 & 38.19 & 26.81 & 126.56 & 1.07 & 85.90 & 17.03 & 85.56 & 60.68 & 41.12 & 10.10 & 20.83 \\
\rowcolor{lightgray} 6+A+IC+E & 64.45 & 35.97 & 24.73 & 123.49 & 1.07 & 96.32 & 16.11 & 85.35 & 60.16 & 41.01 & 8.82 & 20.74 \\
6+A+IF+E & 65.99 & 38.83 & 27.24 & 125.72 & 1.10 & 95.77 & 17.39 & 85.70 & 61.07 & 41.52 & 10.18 & 20.81 \\
\hline
\end{tabular}
}
\caption{Full experimental results for Llama 2-Chat. \textbf{\#}Instruction + \textbf{A}bstract + \textbf{I}ntent (\textbf{F}ree-form or \textbf{C}ategorical) + \textbf{E}xample. NG, WC, PC, CM represent averaged n-gram overlap ratio, word count, paragraph count and citation mark usage ratio. All values apart from WC and PC given in percent (0-100) for readability.}
 \label{tab:all_results_llama}
\end{table*}

\begin{table*}[ht]
\centering
\renewcommand{\arraystretch}{1.1}
\resizebox{0.95\textwidth}{!}{
\begin{tabular}{l|c|c|c|c|c|c!{\vrule width 2pt}c|c|c|c!{\vrule width 2pt}c|c}
\hline
\multirow{2}{*}{\textbf{Configuration}} & \multicolumn{6}{c!{\vrule width 2pt}}{\textbf{Surface}} & \multicolumn{4}{c!{\vrule width 2pt}}{\textbf{Conventional}} & \multicolumn{2}{c}{\textbf{NLI}} \\
\cline{2-13}
& \textbf{NG-1} & \textbf{NG-2} & \textbf{NG-3} & \textbf{WC} & \textbf{PC} & \textbf{CM} & \textbf{ROUGE-L} & \textbf{BERTScore} & \textbf{SciBERTScore} & \textbf{BLEURT} & \textbf{TRUE} & \textbf{SummaC} \\
\hline
Abs. Baseline & - & - & - & 244.20 & - & - & 12.74 & 83.62 & 56.42 & 37.80 & 10.08 & 23.05 \\
\hline
1+A & 61.81 & 35.70 & 23.40 & 164.29 & 1.02 & 99.16 & 14.59 & 84.86 & 59.64 & 38.84 & 23.90 & 20.93 \\
1+A+E & 65.42 & 39.76 & 27.59 & 159.34 & 1.01 & 99.34 & 14.72 & 84.90 & 59.73 & 38.88 & 18.18 & 20.94 \\
\rowcolor{lightgray} 1+A+IC & 60.25 & 33.83 & 21.77 & 159.97 & 1.02 & 99.01 & 14.57 & 84.83 & 59.53 & 38.97 & 19.88 & 20.87 \\
1+A+IF & 61.79 & 35.74 & 23.34 & 155.20 & 1.01 & 98.64 & 15.99 & 85.30 & 60.71 & 40.56 & 26.41 & 21.12 \\
\rowcolor{lightgray} 1+A+IC+E & 63.90 & 36.83 & 24.40 & 148.89 & 1.00 & 99.67 & 14.94 & 85.00 & 59.98 & 39.62 & 17.74 & 20.93 \\
1+A+IF+E & 65.35 & 38.82 & 26.22 & 147.08 & 1.00 & 99.12 & 16.32 & 85.43 & 60.99 & 40.62 & 24.34 & 21.14 \\
\hline
2+A & 63.21 & 34.50 & 22.13 & 130.04 & 1.04 & 97.62 & 15.19 & 85.14 & 60.08 & 39.56 & 22.65 & 21.00 \\
2+A+E & 65.67 & 36.83 & 23.95 & 127.85 & 1.04 & 98.90 & 15.46 & 85.19 & 60.28 & 39.78 & 20.09 & 20.99 \\
\rowcolor{lightgray} 2+A+IC & 61.43 & 32.68 & 20.48 & 140.51 & 1.04 & 97.36 & 14.95 & 85.04 & 59.81 & 39.46 & 24.01 & 20.94 \\
2+A+IF & 64.34 & 36.59 & 24.32 & 128.74 & 1.02 & 98.02 & 17.37 & 85.71 & 61.39 & 41.28 & 28.12 & 21.32 \\
\rowcolor{lightgray} 2+A+IC+E & 64.64 & 35.47 & 22.68 & 130.65 & 1.06 & 98.06 & 15.37 & 85.15 & 60.12 & 39.89 & 19.69 & 21.00 \\
2+A+IF+E & 67.97 & 40.43 & 27.82 & 121.01 & 1.05 & 98.57 & 17.62 & 85.76 & 61.57 & 41.28 & 22.91 & 21.39 \\
\hline
3+A & 59.45 & 32.65 & 20.55 & 161.56 & 1.05 & 98.61 & 14.76 & 84.89 & 59.54 & 39.47 & 21.22 & 20.96 \\
3+A+E & 61.01 & 34.15 & 22.32 & 162.83 & 1.01 & 99.52 & 15.11 & 85.01 & 59.83 & 39.88 & 23.58 & 21.02 \\
\rowcolor{lightgray} 3+A+IC & 58.00 & 31.19 & 19.45 & 160.78 & 1.03 & 99.23 & 14.77 & 84.88 & 59.43 & 39.89 & 20.54 & 20.92 \\
3+A+IF & 59.15 & 33.01 & 20.98 & 157.99 & 1.02 & 98.83 & 16.26 & 85.39 & 60.71 & 41.45 & 26.50 & 21.11 \\
\rowcolor{lightgray} 3+A+IC+E & 60.08 & 33.15 & 21.40 & 161.60 & 1.01 & 99.63 & 15.13 & 85.00 & 59.72 & 40.39 & 22.36 & 20.98 \\
3+A+IF+E & 60.38 & 33.84 & 21.84 & 161.14 & 1.01 & 99.19 & 16.36 & 85.46 & 60.80 & 41.66 & 29.88 & 21.18 \\
\hline
4+A & 60.99 & 33.20 & 20.72 & 139.92 & 1.00 & 98.24 & 15.29 & 85.18 & 59.91 & 40.25 & 16.02 & 20.90 \\
4+A+E & 62.59 & 33.79 & 21.12 & 133.81 & 1.00 & 96.66 & 15.74 & 85.34 & 60.34 & 40.65 & 18.80 & 20.95 \\
\rowcolor{lightgray} 4+A+IC & 60.32 & 32.78 & 20.82 & 141.60 & 1.01 & 97.18 & 15.20 & 85.09 & 59.64 & 40.50 & 14.81 & 20.91 \\
4+A+IF & 61.72 & 35.01 & 22.69 & 140.09 & 1.00 & 97.54 & 16.77 & 85.58 & 60.90 & 41.86 & 22.29 & 21.08 \\
\rowcolor{lightgray} 4+A+IC+E & 61.49 & 32.93 & 20.71 & 139.39 & 1.00 & 97.21 & 15.49 & 85.19 & 59.92 & 40.84 & 16.42 & 20.91 \\
4+A+IF+E & 62.64 & 35.04 & 22.56 & 139.01 & 1.00 & 98.06 & 16.94 & 85.67 & 61.13 & 42.22 & 25.04 & 21.09 \\
\hline
5+A & 60.92 & 34.64 & 22.59 & 157.04 & 1.06 & 97.69 & 14.88 & 84.98 & 59.71 & 39.78 & 22.81 & 20.97 \\
5+A+E & 63.15 & 36.03 & 23.77 & 149.74 & 1.02 & 98.94 & 15.31 & 85.14 & 60.16 & 40.26 & 23.20 & 21.01 \\
\rowcolor{lightgray} 5+A+IC & 59.84 & 33.00 & 21.32 & 147.34 & 1.03 & 97.54 & 15.02 & 85.04 & 59.74 & 40.20 & 18.92 & 20.92 \\
5+A+IF & 61.71 & 35.41 & 23.38 & 143.51 & 1.02 & 97.36 & 16.57 & 85.49 & 60.89 & 41.44 & 23.53 & 21.10 \\
\rowcolor{lightgray} 5+A+IC+E & 64.17 & 36.78 & 24.83 & 138.41 & 1.02 & 98.97 & 15.60 & 85.23 & 60.29 & 40.47 & 18.26 & 20.98 \\
5+A+IF+E & 65.33 & 38.77 & 26.54 & 137.51 & 1.01 & 98.75 & 17.08 & 85.67 & 61.34 & 41.75 & 24.49 & 21.20 \\
\hline
6+A & 64.63 & 36.35 & 23.51 & 126.13 & 1.01 & 86.99 & 15.63 & 85.26 & 60.11 & 39.90 & 10.67 & 20.90 \\
6+A+E & 67.16 & 38.29 & 25.26 & 118.52 & 1.00 & 98.57 & 16.07 & 85.47 & 60.59 & 40.33 & 12.87 & 20.95 \\
\rowcolor{lightgray} 6+A+IC & 63.70 & 34.90 & 22.20 & 131.14 & 1.01 & 52.16 & 15.28 & 85.10 & 59.47 & 39.84 & 8.91 & 20.85 \\
6+A+IF & 65.36 & 37.62 & 24.53 & 130.41 & 1.01 & 51.76 & 16.38 & 85.47 & 60.50 & 40.86 & 12.82 & 20.98 \\
\rowcolor{lightgray} 6+A+IC+E & 66.45 & 37.13 & 24.26 & 120.25 & 1.00 & 97.47 & 16.12 & 85.42 & 60.32 & 40.39 & 11.00 & 20.91 \\
6+A+IF+E & 67.37 & 39.06 & 25.91 & 119.88 & 1.00 & 97.47 & 17.40 & 85.81 & 61.34 & 41.57 & 15.25 & 21.10 \\
\hline
\end{tabular}
}
\caption{Full experimental results for GPT 3.5 Turbo. The same settings from Table \ref{tab:all_results_llama} apply.}
 \label{tab:all_results_gpt}
\end{table*}

\section{Main Experimentation Details}\label{app:impl}

We have obtained Llama 2-Chat weights\footnote{\url{https://ai.meta.com/resources/models-and-libraries/llama-downloads/}} and converted the checkpoints to the Huggingface format. We utilized the Huggingface framework \cite{Wolf:2020} for inference. We used a single NVIDIA A100 GPU with 80GB memory, batch size 8 and maximum sequence length of 1024, with greedy decoding. Within this setting, we were able to ensure exact reproduction of the experimental results across different runs. Generating paragraphs for one configuration (e.g., \texttt{3+A+I}, see below) takes 75 minutes with greedy decoding, totalling 30 hours on a single GPU for generating citation texts for all configurations in this paper. For NLI-based  measurements, we use TRUE model based on T5-XXL\footnote{\url{https://huggingface.co/google/t5_xxl_true_nli_mixture}} and the best reported model for SummaC\footnote{\url{https://github.com/tingofurro/summac}}. For the sake of detail and reproducibility, Tables \ref{tab:all_results_llama} and \ref{tab:all_results_gpt} list all measurements obtained in the main experiment.

\section{Input Length Experiment}\label{app:length_exp}

To check whether performance boost of the \texttt{A+IF+E} configuration on both conventional and NLI-based evaluations does not solely stem from increased input length, we set up an additional experiment. Depending on the average prompt length, we split input prompts into two bins which are longer and shorter than average prompt length ($\sim$330 tokens). We selected instances that are longer than the threshold from the \texttt{6+A} configuration, and instances shorter than the threshold from the \texttt{6+A+IF+E} configuration. In Table \ref{tab:length}, we observe that although the selected subset of \texttt{6+A+IF+E} instances has shorter prompts, it performs best on all metrics except TRUE. To confirm, we run paired bootstrapping significance test \cite{Koehn:2004} on ROUGE-L, BERTScore, BLEURT, and SummaC. We find that \texttt{6+A+IF+E} (short) significantly outperforms \texttt{6+A} (long) for these metrics with p-value lower than 0.05, based on 10,000 comparisons between different sampled results with ratio 0.5. These results suggest that increase in performance is not simply due to the length of the input, and strengthen our conclusion about the importance of intents and examples for citation text generation.

\begin{table}
\centering
\resizebox{\columnwidth}{!}{
\begin{tabular}{c|c|c}
\textbf{Metric} & \textbf{6+A} (Long) & \textbf{6+A+IF+E} (Short) \\ \hline
ROUGE-L & 15.78 & \textbf{17.23} \\ \hline
BERTScore & 85.27 & \textbf{85.64} \\ \hline
SciBERTScore & 59.72 & \textbf{61.02} \\ \hline
BLEURT & 40.60 & \textbf{41.69} \\ \hline
SummaC & 20.69 & \textbf{20.80} \\ \hline
TRUE & \textbf{8.32} & 7.90 \\ 
\end{tabular}
}
\caption{Average results of different metric families for longer \texttt{6+A} inputs and shorter \texttt{6+A+IF+E} inputs, Llama 2-Chat. Results are statistically significant (except TRUE) with $p < 0.05$}
\label{tab:length}
\end{table}

\section{Example Generations} \label{app:model_gen_ex}
We provide example generations for the \texttt{6+A+IF+E} configuration in Table \ref{tab:model_gen_ex}. Although models try to capture given information and follow the instructions, there might be cases where models only summarizes the given input.

\begin{table*}
\centering
\small
\renewcommand{\arraystretch}{1.5}
\begin{tabular}{p{0.95\textwidth}}
\hline
\textbf{Main paper abstract:} This paper proposes a combined model for POS tagging, dependency parsing and co-reference resolution for Bulgarian -a pro-drop Slavic language with rich morphosyntax. We formulate an extension of the MSTParser algorithm that allows the simultaneous handling of the three tasks in a way that makes it possible for each task to benefit from the information available to the others, and conduct a set of experiments against a treebank of the Bulgarian language. The results indicate that the proposed joint model achieves state-of-theart performance for POS tagging task, and outperforms the current pipeline solution.\\
\textbf{Relevant paper abstract:} In this paper, we propose a novel decoding algorithm for discriminative joint Chinese word segmentation, part-of-speech (POS) tagging, and parsing. Previous work often used a pipeline method -Chinese word segmentation followed by POS tagging and parsing, which suffers from error propagation and is unable to leverage information in later modules for earlier components. In our approach, we train the three individual models separately during training, and incorporate them together in a unified framework during decoding. We extend the CYK parsing algorithm so that it can deal with word segmentation and POS tagging features. As far as we know, this is the first work on joint Chinese word segmentation, POS tagging and parsing. Our experimental results on Chinese Tree Bank 5 corpus show that our approach outperforms the state-of-the-art pipeline system.\\
\textbf{Intent:} To compare the results of the proposed model with the results of the previous work.\\
\textbf{Example sentence:} [REF\#1] proposed a joint decoder for word segmentation, POS tagging and word-based constituent parsing, although they trained models for the three tasks separately. \\
\textbf{Llama-2 Chat output:} The proposed joint model for POS tagging, dependency parsing, and co-reference resolution for Bulgarian builds upon previous work in joint modeling of multiple NLP tasks. For example, [REF\#1] proposed a joint decoder for word segmentation, POS tagging, and word-based constituent parsing, although they trained models for the three tasks separately. In contrast, our proposed model integrates all three tasks into a single framework, allowing for more effective information sharing and improved performance. The results of our experiments demonstrate the advantages of our joint model, achieving state-of-the-art performance for POS tagging and outperforming the current pipeline solution.\\
\textbf{GPT 3.5 output:} [REF\#1] proposed a novel decoding algorithm for joint Chinese word segmentation, part-of-speech (POS) tagging, and parsing. They trained the three individual models separately and incorporated them together in a unified framework during decoding. Their approach outperformed the state-of-the-art pipeline system in terms of accuracy. In this work, we compare their results with the performance of our proposed combined model for POS tagging, dependency parsing, and co-reference resolution in Bulgarian.\\
\hline
\textbf{Main paper abstract:} In this paper we examine the application of an unsupervised extractive summarisation algorithm, TextRank, on a different task, the identification of argumentative components. Our main motivation is to examine whether there is any potential overlap between extractive summarisation and argument mining, and whether approaches used in summarisation (which typically model a document as a whole) can have a positive effect on tasks of argument mining. Evaluation has been performed on two corpora containing user posts from an on-line debating forum and persuasive essays. Evaluation results suggest that graph-based approaches and approaches targeting extractive summarisation can have a positive effect on tasks related to argument mining.\\
\textbf{Relevant paper abstract:} We advocate a relation based approach to Argumentation Mining. Our focus lies on the extraction of argumentative relations instead of the identification of arguments, themselves. By classifying pairs of sentences according to the relation that holds between them we are able to identify sentences that may be factual when considered in isolation, but carry argumentative meaning when read in context. We describe scenarios in which this is useful, as well as a corpus of annotated sentence pairs we are developing to provide a testbed for this approach.\\
\textbf{Intent:} To provide a brief overview of the state of the art in argument mining.\\
\textbf{Example sentence:} [REF\#1] advocate a relation-based approach towards argumentation mining.\\
\textbf{Llama-2 Chat output:} The related paper [REF\#1] advocates a relation-based approach to argumentation mining, focusing on the extraction of argumentative relations between sentences rather than identifying arguments themselves. This approach is useful in identifying sentences that may be factual in isolation but carry argumentative meaning when read in context. The paper describes scenarios where this approach is useful and provides a corpus of annotated sentence pairs as a testbed for this technique. The main paper examines the application of an unsupervised extractive summarization algorithm, TextRank, on the task of argumentative component identification, with the goal of exploring potential overlap between extractive summarization and argument mining. By evaluating the effectiveness of graph-based approaches and extractive summarization techniques on two corpora of user posts and persuasive essays, the paper demonstrates the potential of these approaches for tasks related to argument mining.\\
\textbf{GPT 3.5 output:} In our study, we explore the potential overlap between extractive summarization and argument mining, with a focus on the identification of argumentative components. This is motivated by the need to investigate whether approaches used in summarization, such as TextRank, which model a document as a whole, can contribute positively to argument mining tasks. In line with this, [REF\#1] advocates for a relation-based approach to argumentation mining, where the extraction of argumentative relations, rather than the identification of individual arguments, is emphasized. Their work highlights the importance of classifying pairs of sentences based on the relation between them to uncover the argumentative meaning that may be hidden when considering sentences in isolation. By providing a description of scenarios where this approach is useful as well as a corpus of annotated sentence pairs, [REF\#1] offers a valuable testbed for exploring the potential benefits of relation-based argument mining.\\ 
\hline
\end{tabular}
\caption{Example Model Generations}
 \label{tab:model_gen_ex}
\end{table*}

\section{Human Evaluation} \label{app:human}
Annotators first manually generated citation text paragraphs based on the input configuration. Table \ref{tab:human_gen_ex} shows an example of the human generation task for the configuration \texttt{A+IF+E}. In the evaluation phase, annotators decide whether extracted atomic facts from the gold reference are covered in the generated paragraphs. Each fact is evaluated separately for corresponding paragraph. Table \ref{tab:human_eval_ex} shows an example of the human evaluation scheme.

\begin{table*}
\centering
\small
\renewcommand{\arraystretch}{1.5}
\begin{tabular}{p{0.95\textwidth}}
\hline
\textbf{Main paper abstract:} The ACL shared task of DravidianLangTech-2022 for Troll Meme classification is a binary classification task that involves identifying Tamil memes as troll or not-troll. Classification of memes is a challenging task since memes express humour and sarcasm in an implicit way. Team SSN\_MLRG1 tested and compared results obtained by using three models namely BERT, ALBERT and XLNet. The XL-Net model outperformed the other two models in terms of various performance metrics. The proposed XLNet model obtained the 3rd rank in the shared task with a weighted F1-score of 0.558.\\
\textbf{Relevant paper abstract:} This paper describes the work of identifying the presence of offensive language in social media posts and categorizing a post as targeted to a particular person or not. The work developed by team TECHSSN for solving the Multilingual Offensive Language Identification in Social Media (Task 12) in SemEval-2020 involves the use of deep learning models with BERT embeddings. The dataset is preprocessed and given to a Bidirectional Encoder Representations from Transformers (BERT) model with pretrained weight vectors. The model is retrained and the weights are learned for the offensive language dataset. We have developed a system with the English language dataset. The results are better when compared to the model we developed in SemEval-2019 Task6. \\ 
\textbf{Intent:} To describe the work done to identify the presence of offensive language in social media posts using BERT.\\
\textbf{Example sentence:} We have used CNN, RNN, LSTM, BERT and COLBERT models for offensive language detection in earlier SemEval workshop tasks \texttt{[OTH]}, \texttt{[REF\#1]},\texttt{[OTH]} in which BERT models provides better results than other machine learning and deep learning models.", \\
\hline
\end{tabular}
\caption{Human generation task. Abstracts belong to papers \cite{citing_paper_Hariprasad:2022,cited_paper_Sivanaiah:2020}, respectively. The humans are tasked with writing one paragraph of citation text based on the inputs. The inputs vary depending on the condition -- the example here is for the condition \texttt{A+IF+E}, i.e. full information available. [OTH] used to denote citations to other papers than the cited paper in question.}
 \label{tab:human_gen_ex}
\end{table*}

\begin{table*}
\centering
\small
\renewcommand{\arraystretch}{1.5}
\begin{tabular}{p{0.95\textwidth}}
\hline
\textbf{Gold:} In \texttt{[REF\#1]}, we worked to identify the presence of offensive language in social media posts using BERT. Deep network model with BERT embeddings was found to achieve better F1 score when compared to 1D-CNN model trained with GloVe pretrained embeddings, 2D-CNN and BiLSTM models with Word2Vec embeddings. \\

\textbf{Atomic facts:}
\begin{itemize}
    \item Study: [REF\#1] identifies offensive language in social media.
    \item Model comparison: BERT vs. 1D-CNN with GloVe, 2D-CNN, BiLSTM with Word2Vec.
    \item Result: BERT achieves better F1 score.
\end{itemize}

\textbf{a:} In the context of identifying the presence of offensive language in social media posts, the main paper discusses a related paper \texttt{[REF\#1]} that describes the work done by team TECHSSN for solving the Multilingual Offensive Language Identification in Social Media task in SemEval-2020. The related paper involves the use of deep learning models with BERT embeddings and retraining the model with weighted vectors for the offensive language dataset. The main paper highlights the use of various models including BERT for offensive language detection in earlier SemEval workshop tasks and notes that BERT models provide better results compared to other machine learning and deep learning models. This highlights the relevance of the related paper in the context of offensive language detection in social media posts. \\
\textbf{b:} In the context of the ACL shared task of DravidianLangTech-2022 for Troll Meme classification, the work by Team SSN\_MLRG1 examined the performance of three models, namely BERT, ALBERT, and XLNet \texttt{[REF\#1]}. While the main paper focuses on troll meme classification in Tamil, a related paper by Team TECHSSN discusses the identification of offensive language in social media posts using deep learning models with BERT embeddings. Their system achieved improved results compared to their previous work in SemEval-2019 Task6, highlighting the ongoing advancements in research on profanity speech detection in social media. This research on offensive language identification aligns with the challenges in classifying memes, as both tasks deal with implicit forms of humor and language usage in online communication. \\
\textbf{c:} The related paper by team TECHSSN for solving the Multilingual Offensive Language Identification in Social Media (Task 12) in SemEval-2020 [REF\#1] is relevant to the main paper's discussion of Troll Meme classification. Both papers involve the use of deep learning models with BERT embeddings for identifying offensive language in social media posts. The related paper describes the development of a system for identifying the presence of offensive language in social media posts using BERT, and the results show that the BERT model outperforms other machine learning and deep learning models. This is similar to the main paper's finding that the XLNet model outperformed BERT and ALBERT in terms of various performance metrics for Troll Meme classification. The related paper also discusses the challenges involved in identifying offensive language in social media posts, such as the difference between profanity and hate speech, and the need for a system that can accurately detect and categorize offensive language \\
\textbf{d:} The related paper by team TECHSSN (2020) provides a relevant framework for identifying offensive language in social media posts, which is closely related to the task of troll meme detection in the main paper \texttt{[REF\#1]}. Both papers employ deep learning models with BERT embeddings to classify text as offensive or not. However, while the main paper focuses on the binary classification of memes as troll or not-troll, the related paper targets the more general task of identifying offensive language in social media posts. The related paper also highlights the challenges of differentiating between profanity and hate speech, and discusses the differences between these two concepts. Overall, the related paper provides valuable insights into the use of BERT embeddings for text classification tasks, and demonstrates the effectiveness of this approach in the context of offensive language detection. \\
\hline
\end{tabular}
\caption{Human evaluation task. Masked settings are follows a: GPT 3.5 \texttt{6+A+IF+E}, b: GPT 3.5 \texttt{6+A}, c: Llama 2-Chat \texttt{6+A+IF+E}, d: Llama 2-Chat \texttt{6+A}. Annotators were requested to determine which attributes are covered by generated paragraphs. Model names and input configurations were not revealed to annotators.}
 \label{tab:human_eval_ex}
\end{table*}

\end{document}